\documentclass[10pt,twocolumn,letterpaper]{article}
\usepackage{cvpr}
\usepackage{times}
\usepackage{epsfig}
\usepackage{graphicx}
\usepackage{amsmath}
\usepackage{amssymb}

\usepackage{multirow}
\usepackage{hhline}
\usepackage{subfig}
\usepackage{wrapfig}
\usepackage{amsmath}
\usepackage{array}
\usepackage{xcolor}

\newcolumntype{C}[1]{>{\centering\arraybackslash}m{#1}}

\cvprfinalcopy 

\def\cvprPaperID{****} 

\ifcvprfinal\fi
\begin{document}

\title{Regularizing Activation Distribution for Training Binarized Deep Networks}

\author{Ruizhou Ding, Ting-Wu Chin, Zeye Liu, Diana Marculescu\\
Carnegie Mellon University\\
{\tt\small \{rding, tingwuc, zeyel, dianam\}@andrew.cmu.edu}
}

\maketitle

\begin{abstract}
   Binarized Neural Networks (BNNs) can significantly reduce the inference latency and energy consumption in resource-constrained devices due to their pure-logical computation and fewer memory accesses. However, training BNNs is difficult since the activation flow encounters degeneration, saturation, and gradient mismatch problems. Prior work alleviates these issues by increasing activation bits and adding floating-point scaling factors, thereby sacrificing BNN's energy efficiency. In this paper, we propose to use distribution loss to explicitly regularize the activation flow, and develop a framework to systematically formulate the loss. Our experiments show that the distribution loss can consistently improve the accuracy of BNNs without losing their energy benefits. Moreover, equipped with the proposed regularization, BNN training is shown to be robust to the selection of hyper-parameters including optimizer and learning rate. 
\end{abstract}

\section{Introduction}\label{sec:intro}

Recent years have witnessed tremendous success of Deep Neural Networks (DNNs) in various applications of image, video, speech, natural language, \textit{etc}~\cite{he2016deep,lu2017knowing}. However, the increased computation workload and memory access count required by DNNs pose a burden on latency-sensitive applications and energy-limited devices. Since latency and energy consumption are highly related to \textit{computation} cost and \textit{memory} access count, there has been a lot of research on reducing these two important design metrics~\cite{ding2017lightnn,sandler2018mobilenetv2,he2018amc,chin2019adascale}.  Binarized Neural Networks (BNNs)~\cite{hubara2016binarized} that constrain the network weights and activations to be $\pm1$ have been proven highly efficient on custom hardware~\cite{zhao2017accelerating}. We also show later in Sec.~\ref{sec:BNN} that a typical block of a BNN can be implemented in hardware with merely a few logical operators including XNOR gates, counters and comparators, and therefore greatly reduce the energy consumption and circuit area, as shown in Table~\ref{table:operator-energy}.

In addition to the computational benefit brought by making the whole network binarized, another benefit of BNNs is the huge reduction of memory footprint due to their 1-bit weights and activations. Prior work on extremely low-bit DNNs~\cite{courbariaux2015binaryconnect,faraone2018syq,lin2016fixed,cai2017deep,tang2017train,ding2018lightening} mainly focuses on few-bit weights and uses more bits for activations, while only a few~\cite{hubara2016binarized,lin2017towards} target 1-bit weights \textit{and} activations. However, reading and writing intermediate results (activations) generate a larger memory footprint than the weights~\cite{mishra2018wrpn}. For example, in the inference phase of a full-precision (32-bit) AlexNet with batch size 32, 92.7\% of the memory footprint is caused by activations, while only 7.3\% is caused by weights~\cite{mishra2018wrpn}. Therefore, the memory footprint of BNNs is significantly reduced due to their binary activations.

\begin{figure}[t]
  \centering
  \includegraphics[width=0.47\textwidth]{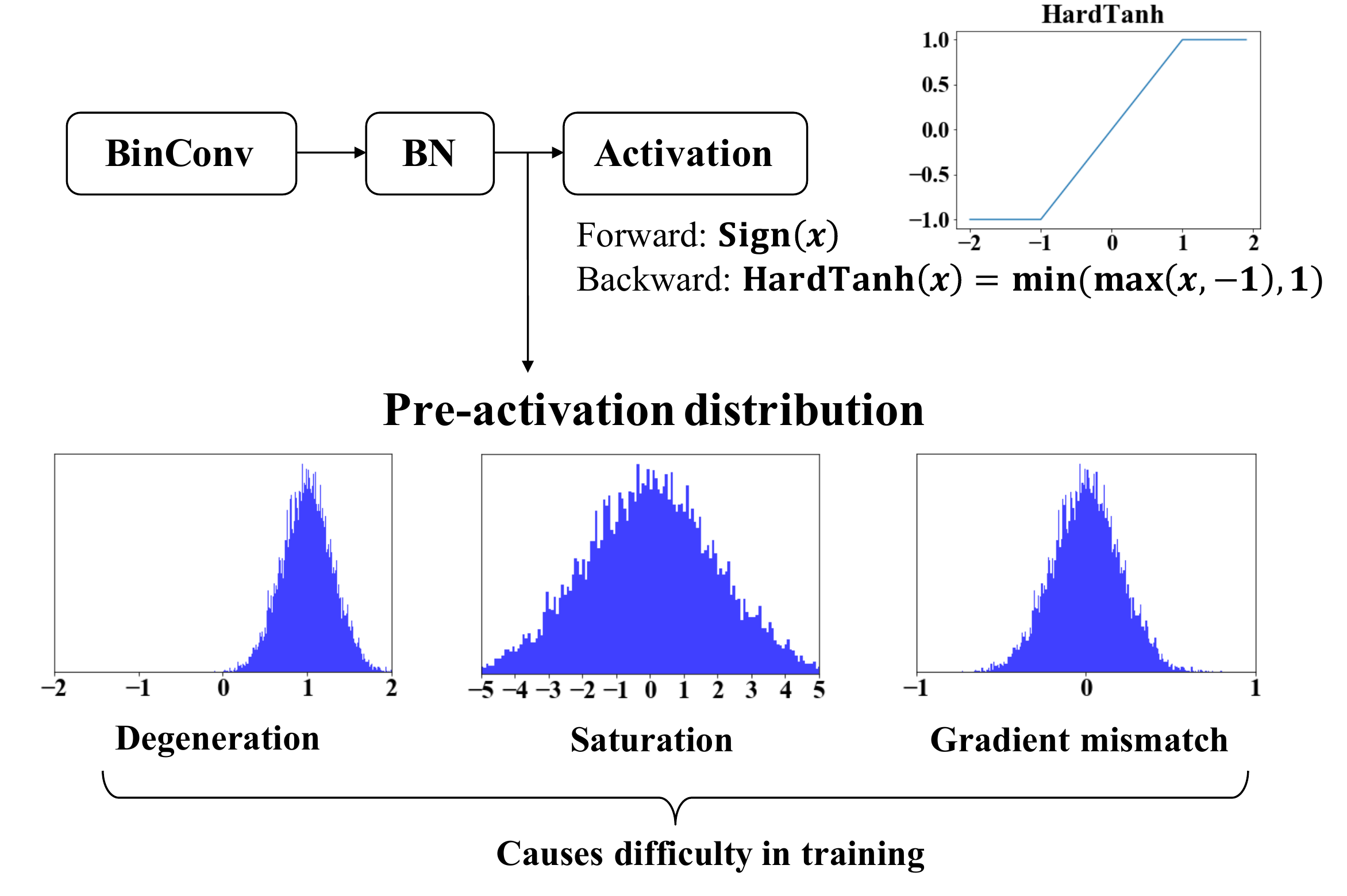}
  \vspace{-6pt}
  \caption{The basic Conv-BN-Act structure for BNN (BinConv: binary convolution; BN: batch normalization). The pre-activation distribution may exhibit from degeneration, saturation or gradient mismatch problem that causes difficulty in training.}\label{fig:hist_problems}
\end{figure}

However, training accurate BNNs requires careful hyper-parameter selection~\cite{alizadeh2018a}, which makes the process more difficult than for their full-precision counterparts. Prior work has shown that this difficulty arises from the bounded activation function and the gradient approximation of the non-differentiable quantization function~\cite{cai2017deep}. Even for full-precision DNNs, bounded activation functions (\textit{e.g.}, \textit{Sigmoid} or \textit{Tanh}) usually lead to lower accuracy compared to the unbounded ones (\textit{e.g.}, \textit{ReLU}, \textit{leaky ReLU}, or \textit{SELU}) due to the gradient vanishing problem~\cite{glorot2010understanding,chen2018understanding}. For binarized networks, a bounded activation (\emph{i.e.}, \textit{Sign} function) is used to lead to binary activations, and the \emph{HardTanh} activation function is commonly used for gradient approximation~\cite{hubara2016binarized,rastegariECCV16,tang2017train}. As shown in Fig.~\ref{fig:hist_problems}, these bounded activation functions bring the following challenges (we use a convolutional layer as an example for illustration purposes): (i) \textit{Degeneration}: If almost all the pre-activations of a channel have the same sign, then this channel will output nearly constant activations. In an extreme case, this channel degenerates to a constant.
(ii) \textit{Saturation}: If most of the pre-activations of a channel have a larger absolute value than the \emph{HardTanh} threshold (\textit{i.e.}, $|a|\ge1$), then the gradients for these pre-activations will be zero. 
(iii) \textit{Gradient mismatch}: If the absolute values of pre-activations are consistently smaller than the threshold (\textit{i.e.}, $|a|<1$), then this is equivalent to using a straight-through estimator (STE) for gradient computation~\cite{bengio2013estimating}. While the STE generally performs well in computing gradients of staircase functions when training fixed-point DNNs, using STE for computing the gradient of \textit{Sign} function causes larger approximation error than staircase function, and therefore causes worse gradient mismatch~\cite{cai2017deep}.

Due to the difficulty of BNN training, prior work along this track has traded the benefit of extremely-low energy consumption for higher accuracy. Hubara \textit{et al.} largely increase the number of filters/neurons per convolutional/fully-connected layer~\cite{hubara2016binarized}. Thus, while a portion of filters/neurons are blocked due to degeneration or gradient saturation, there is still a large \textit{absolute} number of filters/neurons that can work well. Similarly, Mishra \textit{et al.} also increase the width of the network to keep the BNN accuracy high~\cite{mishra2018wrpn}. 

In addition to increasing the number of network parameters, lots of work sacrifices BNNs' pure-logical advantage by relaxing the precision constraint. Rastegari \textit{et al.} approximate a full-precision convolution by using a binary convolution followed by a floating-point element-wise multiplication with a scaling matrix.~\cite{rastegariECCV16}. Tang \textit{et al.} use multiple-bit binarization for activations, which requires floating-point operators to compute the mean and residual error of activations~\cite{tang2017train}. Lin \textit{et al.} approximate each filter and activation map using a weighted sum of multiple binary tensors~\cite{lin2017towards}. All these approaches use scaling factors for weights and activations, making fixed-point multiplication and addition necessary for hardware implementation. Liu \textit{et al.} added skip connections with floating point computations to the model~\cite{liu2018bi}. While the models resulting from these approaches use XNOR convolution kernels, the extra multiplications and additions are not negligible. As shown in Table.~\ref{table:layer-energy}, the energy cost for a typical convolutional layer of BNN is lower than the other binarized DNNs. The layer setting and the proposed approach for energy cost estimation are introduced in Appendix~\ref{appendix:energy-cost}. Furthermore, since the hardware implementation of BNNs do not require digital signal processing (DSP) units, they greatly save circuit area and thus, can benefit IoT applications that have stringent area constraint~\cite{gobieski2018intelligence,conti2017iot}.

\begin{table}[t]
\begin{center}
\caption{Computational energy consumption and circuit area for different computation operators using a commercial 65nm process design kit~\cite{STlib}. The multiplier and adder are both 16-bit fixed-point operators.}
\label{table:operator-energy}
\vspace{-12pt}
\begin{tabular}{C{1.6cm}|C{1.15cm}C{1.15cm}|C{1.15cm}C{1.15cm}}
\hline
& Energy (pJ) & Relative cost  & Area ($\mu$m$^2$) & Relative cost  \\
\hline
XNOR & 7.6$\times$10$^{-4}$ & 1$\times$ & 4.2 & 1$\times$ \\
Counter & 7.8$\times$10$^{-4}$ & 10$\times$ & 52 & 12$\times$ \\
Comparator & 1.1$\times$10$^{-2}$ & 14$\times$ & 52 & 12$\times$ \\
Multiplier & 1.6 & 2109$\times$ & 3.0$\times$10$^3$ & 718$\times$ \\
Adder & 4.8$\times$10$^{-2}$& 64$\times$ & 1.6$\times$10$^2$ & 37$\times$\\
\hline 
\end{tabular}
\vspace{-12pt}
\end{center}
\end{table}

\begin{table}[t]
\begin{center}
\caption{Computational energy for a convolutional layer with different types of binarizations.}
\label{table:layer-energy}
\vspace{-9pt}
\begin{tabular}{C{2.5cm}C{1cm}C{1cm}C{1cm}}
\hline
& Pure-logical & Energy ($\mu$J) & Relative cost  \\
\hline
BNN~\cite{hubara2016binarized} & Yes & 1.42 & 1$\times$ \\
XNOR-Net~\cite{rastegariECCV16} & No & 4.34 & 3$\times$ \\
ABC-Net~\cite{lin2017towards} & No & 24.6 & 17$\times$ \\
\hline
\end{tabular}
\vspace{-12pt}
\end{center}
\end{table}

In this paper, we propose a general framework for activation regularization to tackle the difficulties encountered during BNN training. While prior work on weight initialization~\cite{glorot2010understanding} and batch normalization~\cite{ioffe2015batch} also regularizes activations, it does not address the challenges mentioned earlier for BNNs, as detailed in Sec.~\ref{sec:exp-regularized}. Instead of regularizing the activation distribution in an implicit fashion as done in prior work~\cite{glorot2010understanding,ioffe2015batch}, we shape the distribution explicitly by embedding the regularization in the loss function. This regularization is shown to effectively alleviate the challenges for BNNs, and consistently increase the accuracy. Specifically, adding the distribution loss can improve the Top-1 accuracy of BNN AlexNet~\cite{hubara2016binarized} on ImageNet from \textbf{36.1\%} to \textbf{41.3\%}, and improve the binarized wide AlexNet~\cite{mishra2018wrpn} from \textbf{48.3\%} to \textbf{53.8\%}.
In summary, this paper has the following key contributions: 

(i) To the best of our knowledge, we are the first to propose a framework for explicit activation regularization for binarized networks that consistently improve the accuracy. 

(ii) Empirical results show that the proposed distribution loss is robust to the selection of training hyper-parameters. Code is available at: https://github.com/ruizhoud/DistributionLoss.

\section{Related Work}
Prior work has proposed various approaches to regularize the activation flow of full-precision DNNs, mainly to address the gradient vanishing or exploding problem. Ioffe \textit{et al.} propose batch normalization to centralize the activation distribution, accelerate training, and achieve higher accuracy~\cite{ioffe2015batch}. Similarly, Huang \textit{et al.} normalize the weights with zero mean and unit norm followed by scaling factors~\cite{huang2017centered}. Shang \textit{et al.} extend the normalization idea to residual networks using normalized propagation~\cite{shang2017exploring}, while Ba \textit{et al.} and Salimans \textit{et al.} normalize the activations of Recurrent Neural Network (RNN) by layer-wise normalization and weight reparameterization, respectively~\cite{ba2016layer,salimans2016weight}. In addition, some prior work develops good initialization strategy to regularize the activations in the initial state~\cite{mishkin2015all,xie2017all}, or proposes new activation functions to maintain stable activation distribution across layers~\cite{klambauer2017self,maas2013rectifier}.

However, these approaches on full-precision networks do not address the difficulty of training networks with binarized activations. Prior work on binarized DNNs alleviates this problem mainly by approximating the full-precision activations with multiple-bit representations and floating-point scaling factors~\cite{tang2017train,cai2017deep,mishra2018wrpn,polino2018model,mishra2018apprentice,faraone2018syq,lin2017towards,hou2017lossaware}. Tang~\textit{et al.} introduced scaling layers and use 2 bits for activations~\cite{tang2017train}. Cai~\textit{et al.} use multi-level activation function for inference and variants of ReLU for gradient computation to reduce gradient mismatch~\cite{cai2017deep}. Polino~\textit{et al.} leverage knowledge distillation to guide training and improve the accuracy with multiple bits for activations~\cite{polino2018model}. Lin~\textit{et al.} approximate both weights and activations with multiple binary bases associated with floating-point coefficients~\cite{lin2017towards}. While these approaches improve the accuracy for binarized networks, they sacrifice the energy efficiency due to the increased bits and the required DSP units for the additions and multiplications.

\section{Activation Regularization}
In this section, we first show that BNN blocks can be implemented with pure-logical operators in hardware while the other binarized networks (including XNOR-Net and ABC-Net) based on scaling factors require additional full-precision operations. Then, we propose a framework to address the problems of activation and gradient flow incurred in the training process of BNNs. Finally, we discuss the effectiveness of this framework. 

\subsection{Binarized DNNs}\label{sec:BNN}
Binarized DNNs constrain the weights and activations to be $\pm 1$, making the convolution between weights and activations use only \emph{xnor} and \emph{count} operators. In this subsection, we introduce the structure of three typical binarized DNNs, and analyze their hardware implication. 

\begin{figure}[t]
  \centering
  \includegraphics[width=0.47\textwidth]{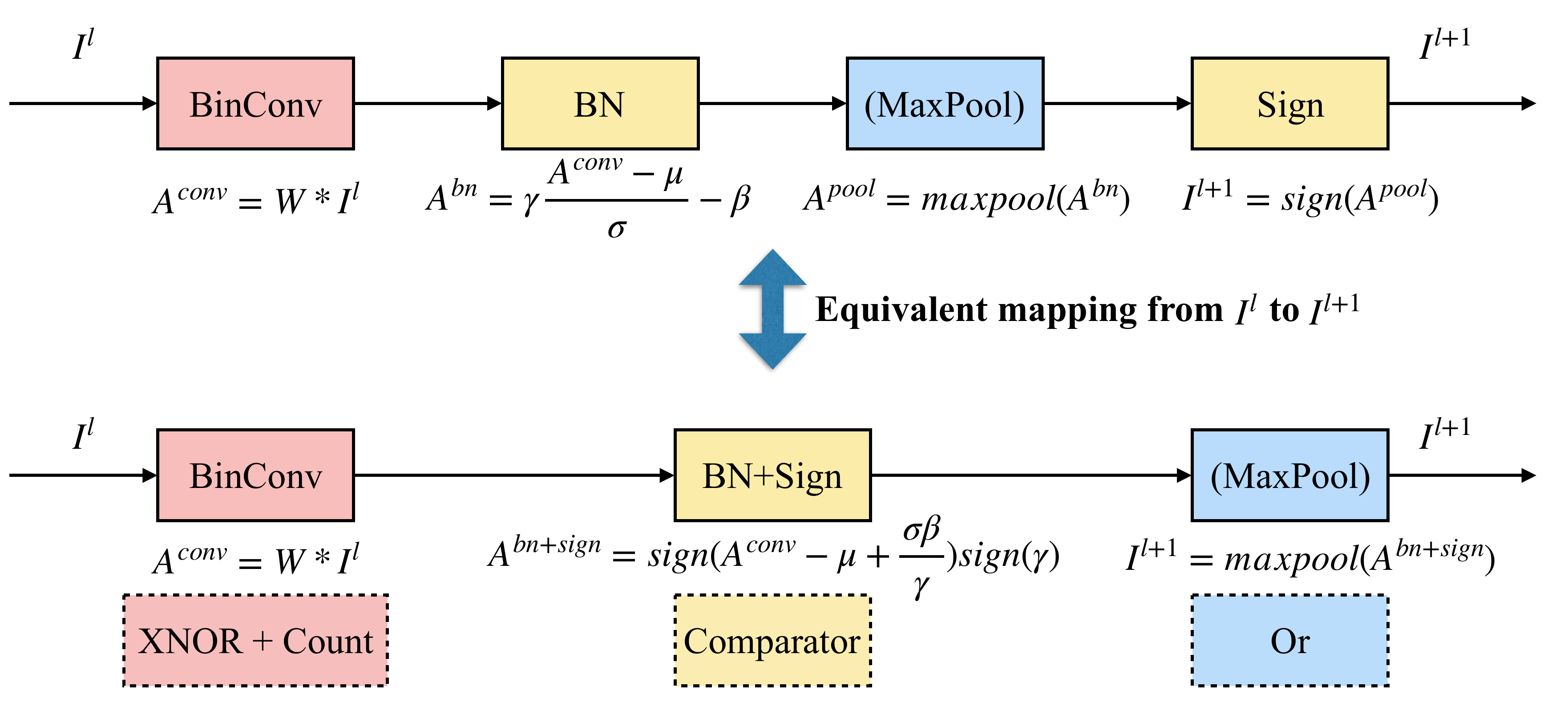}
  \caption{Basic block for convolutional BNN~\cite{hubara2016binarized}. The activations $I^l$ and weights $W$ are binarized to $\pm1$. The inference of this block can be implemented on hardware with only logical operators.}\label{fig:bnn_block}
\end{figure}

\textbf{\textit{BNN: }}
As shown in Fig~\ref{fig:bnn_block}, the basic block for BNN~\cite{hubara2016binarized} is composed of a binary convolution, a batch normalization (BN) and an optional max pooling layer, followed by a sign activation function. Without changing the input-output mapping of this block, we can reorder the max pooling layer and the sign function, and then, combine the BN layer and sign function to be a comparator of the convolution results $A^{conv}$ and input-independent variables $\mu+\frac{\sigma\beta}{\gamma}$, where $\mu$ and $\sigma$ are the moving mean and variance of per-channel activations, which are obtained from training data and fixed in the testing phase; $\beta$ and $\gamma$ are trainable parameters in the BN layer. Therefore, the inference of this BNN block can be implemented in hardware with pure-logical operators. This transformation can also be applied to binarized fully-connected layers followed by BN, pooling and sign function.

\textbf{\textit{XNOR-Net: }}
Different from BNN, XNOR-Net~\cite{rastegariECCV16} approximates the activations $A^{bn}$ after the BN layer with their signs and scaling factors computed by the average of the absolute values of these activations, as shown in Fig.~\ref{fig:xnor_block} in Appendix. Since the scaling factors are input-dependent, the full-precision multiplications and additions cannot be eliminated.

\textbf{\textit{ABC-Net: }}
ABC-Net~\cite{lin2017towards}, shown in Fig.~\ref{fig:abc_block} in Appendix, approximates both weights and activations with a linear combination of pre-defined bases, and therefore making the convolution kernel binarized. However, the approximation prior to binary convolution and the scaling operations after the convolution require full-precision multiplications and additions that cannot be eliminated.

\subsection{Regularizing Activation Distribution}
In this section we first introduce some notations and formally define the difficulties encountered when training BNNs. We denote $A^{b,l,c}$ as the pre-activations (activations prior to the \textit{Sign} function) for the $c$-th channel of the $l$-th layer for the $b$-th batch of data. Thus, $A^{b,l,c}$ is a 3D tensor with size $B\times W\times H$ where $B$ is the batch size, $W$ and $H$ are the width and height of the activation map. From this point on, to avoid clutter, we will omit the superscript of $A$ whenever possible. $A_{(q)}$ denotes the $q$ quantile of $A$'s elements where $0\le q\le 1$. We define degeneration, saturation, and gradient mismatch as follows:
\begin{equation}
    \begin{split}
        & \textbf{Degeneration: } A_{(0)}\ge0~\text{or}~A_{(1)}\le0\\
        & \textbf{Saturation: } |A|_{(0)}\ge 1 \\ 
        & \textbf{Gradient mismatch: } |A|_{(1)}\le 1\\
    \end{split}
\end{equation}
where $|A|_{(q)}$ is the $q$ quantile for $|A|$, and we use 1 because it is the threshold of \textit{HardTanh} activation shown in Fig.~\ref{fig:hist_problems}.

\begin{figure}[t]
  \centering
  \includegraphics[width=0.35\textwidth]{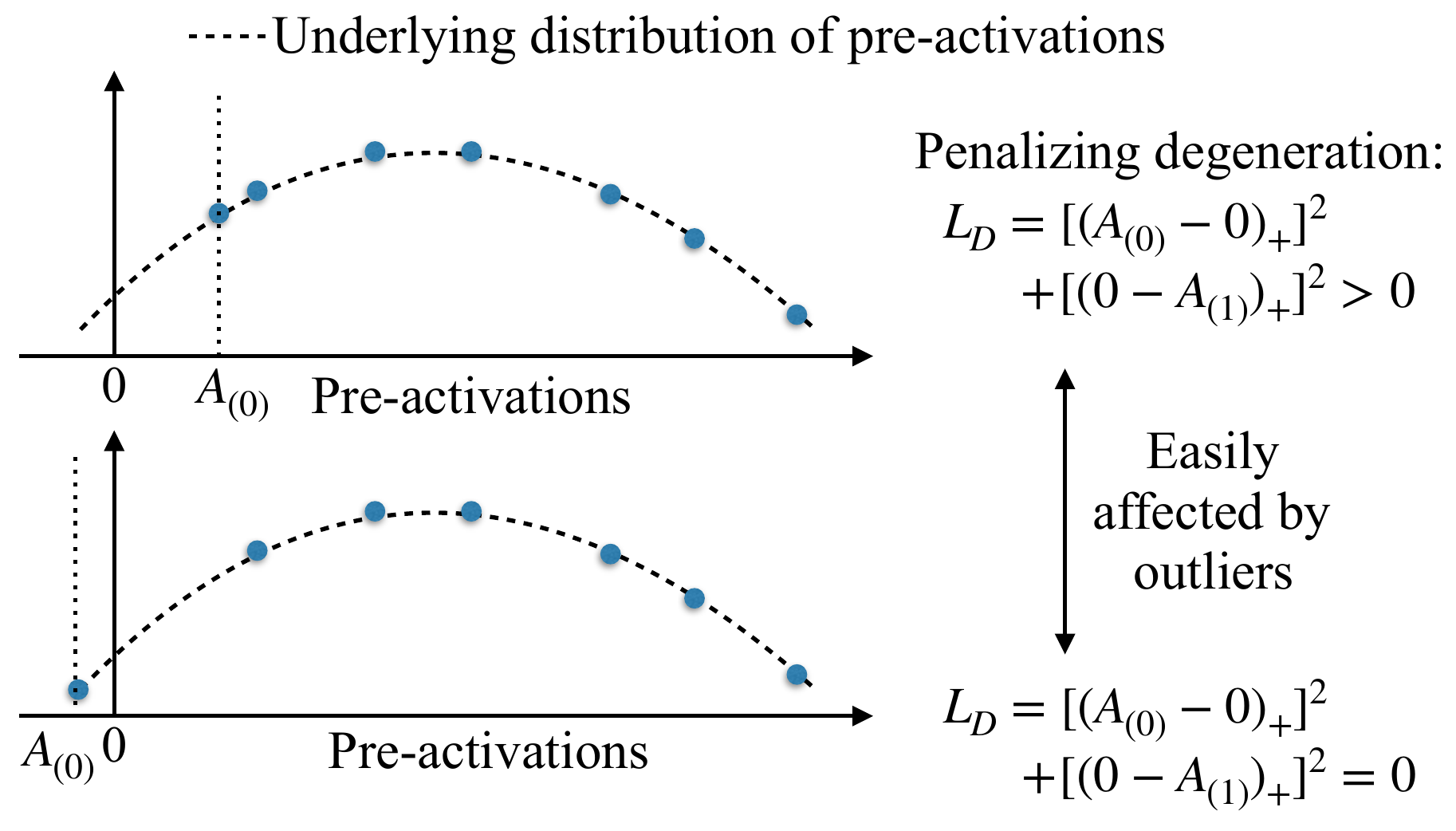}
  \caption{Motivation for adjusting regularization. The loss function directly formulated from hypothesis (\textit{e.g.}, degeneration) relies on the minimum (or maximum) of the pre-activations, and therefore is sensitive to outliers.}\label{fig:motivation-AR}
\end{figure}

To alleviate the aforementioned problems, we propose to add the distribution loss in the objective function to regularize the activation distribution. Using degeneration as an example, an intuitive way of formulating a loss to avoid the degeneration problem for $A$ is $L_{D}=[(A_{(0)}-0)_+]^2+[(0-A_{(1)})_+]^2$, where $(.)_+$ is the ReLU function. However, this may lead to too loose regularization since a small outlier can make this loss zero, as shown in Fig.~\ref{fig:motivation-AR}. In addition, this formulation of $L_{D}$ is not differentiable $w.r.t.$ the pre-activations $A$.

\begin{figure}[t]
  \centering
  \includegraphics[width=0.47\textwidth]{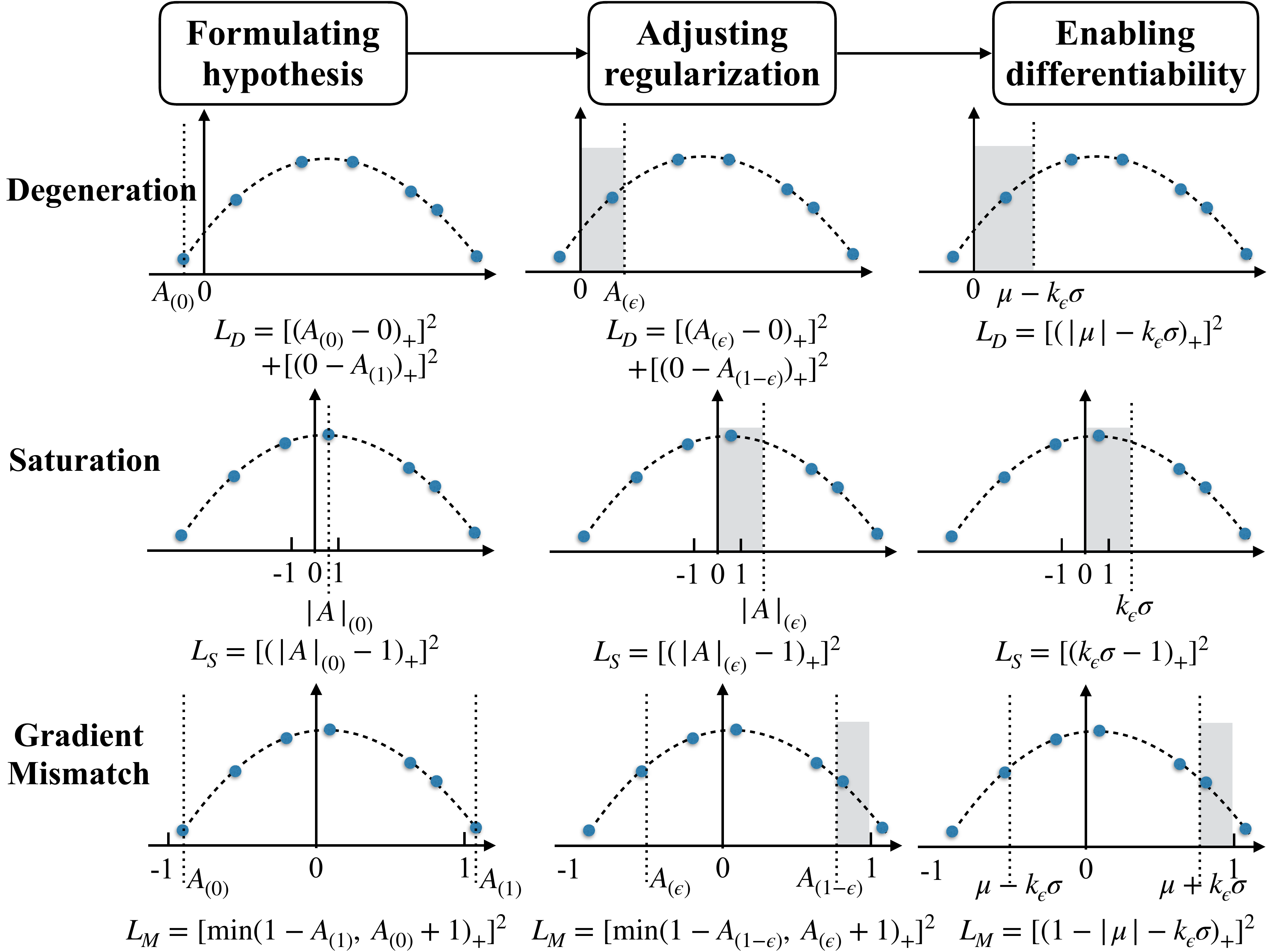}
  \caption{Proposed framework for formulating the differentiable loss function to regularize activation distribution. Starting from the three hypotheses (``degeneration", ``saturation" and ``gradient mismatch"), we can formulate the loss function $L_D$, $L_S$ and $L_M$ for them, respectively. We omit the superscript for $A$ and $L$ for better representation.}\label{fig:method}
\end{figure}

Therefore, we propose a three-stage framework consisting of hypothesis formulation, adjusting regularization, and enabling differentiability, to systematically formulate an outlier-robust and differentiable regularization, as shown in Fig.~\ref{fig:method}. First, based on the prior hypothesis about the activation distribution, we can formulate a loss function to penalize the unwanted distribution. Then, if this formulation uses large-variance estimators (\textit{e.g.}, maximum or minimum of samples), we can use relaxed estimators (\textit{e.g.}, quantiles) to increase robustness to outliers. Finally, if the formulated loss is not differentiable, we need to approximate it by assuming the type of parametric distribution (\textit{e.g.}, Gaussian), and approximate the non-differentiable estimators with the distribution parameters. 

\vspace{-12pt}
\paragraph{Degeneration.}
We first formulate the degeneration hypothesis in the loss function as $L_{D}=[(A_{(0)}-0)_+]^2+[(0-A_{(1)})_+]^2$. To make the loss function more robust to outliers, we adjust the regularization by using relaxed quantiles $\epsilon$ and $1-\epsilon$, with $L_{D}=[(A_{(\epsilon)}-0)_+]^2+[(0-A_{(1-\epsilon)})_+]^2$. Then, to make $L_{D}$ differentiable so that it can fit in the backpropagation training, we first assume a parameterized distribution for the pre-activations $A$ and then use its parameters to formulate a differentiable $L_{D}$. Based on the heuristics from prior art~\cite{lin2016fixed,cai2017deep,park2017weighted}, we assume that the values of $A$ follow a Gaussian distribution $\mathcal{N}(\mu, (\sigma)^2)$, where the $\mu$ and $\sigma$ can be estimated by the sample mean and standard deviation over the 3D tensor. Thus, we can formulate the $\epsilon$ quantile by $\mu-k_\epsilon\sigma$ where $k_\epsilon$ is a constant determined by $\epsilon$. Therefore, $L_{D}=[(\mu-k_\epsilon\sigma-0)_+]^2+[(0-(\mu+k_\epsilon\sigma))_+]^2 = [(|\mu|-k_\epsilon\sigma)_+]^2$.

\vspace{-12pt}
\paragraph{Saturation.}
The saturation problem can be penalized by $L_S=[(|A|_{(0)}-1)_+]^2$, where $|A|_{(0)}$ is the minimum value of $|A|$. By adjusting the regularization, we have $L_S=[(|A|_{(\epsilon)}-1)_+]^2$. Since $L_D$ already eliminates the degeneration problem, we find that simply assuming $A$ has a zero mean (\textit{i.e.}, $\mathcal{N}(0, (\sigma)^2)$) works well empirically. Thus, the loss function is formulated as $L_S=[(k_\epsilon\sigma-1)_+]^2$.

\vspace{-12pt}
\paragraph{Gradient mismatch.}
When most of the activations lie in the range of [-1,1], the backward pass is simply using a STE for the gradient computation of the sign function, causing the gradient mismatch problem. Therefore, we can formulate the loss as $L_{M}=[\min(1-A_{(1)},~A_{(0)}+1)_+]^2$. Similarly, relaxing the regularization leads to $L_{M}=[\min(1-A_{(1-\epsilon)},~A_{(\epsilon)}+1)_+]^2$. With a Gaussian assumption, we have $L_{M}=[\min(1-\mu-k_\epsilon\sigma,~\mu-k_\epsilon\sigma+1)_+]^2=[(1-|\mu|-k_\epsilon\sigma)_+]^2$.

Then, in the training phase, we add the distribution loss for the $b$-th batch of input data: 
\begin{equation}
   L^b_{DL}=\sum_{l,c} L^{b,l,c}_{DL}=\sum_{l,c}L^{b,l,c}_D+L^{b,l,c}_S+L^{b,l,c}_M,
\end{equation}
and the total loss for $b$-th batch is:
\begin{equation}
    L^b_{total}=L^b_{CE}+\lambda L^b_{DL}
\end{equation}
where $L^b_{CE}$ is the cross-entropy loss, and $\lambda$ is a coefficient to balance the losses. 

\subsection{Intuition for the Proposed Distribution Loss}

The distribution loss is proposed to alleviate training problems for pure-logical binarized networks. In contrast with full-precision networks, BNNs use a bounded activation function and therefore exhibit the gradient saturation and mismatch problems. By regularizing the activations, the distribution loss maintains the effectiveness of the back-propagation algorithm, and thus, can speedup training and improve the accuracy. 

Since the distribution loss changes the optimization objective, one concern may be that it will lead to a configuration far from the global optimal of the cross-entropy loss function. However, prior theoretical~\cite{soudry2016no,kawaguchi2016deep,choromanska2015loss} and empirical~\cite{im2016empirical} work has shown that a deep neural network can have many high-quality local optima. Kawaguchi proved that under certain conditions, every local minimum is a global minimum~\cite{kawaguchi2016deep}. Through experiments, Im \textit{et al.} show that using different optimizers, the achieved local optima are very different~\cite{im2016empirical}. These insights show that adding the distribution loss may deviate the training away from the original optimal, but can still lead to a new optimal with high accuracy. Moreover, the distribution loss diagnoses the poor conditions of the activation flow, and therefore may achieve higher accuracy. Our experiment results confirm this hypothesis. 

\section{Experimental Results}
In this section, we first evaluate the accuracy improvement by the proposed distribution loss on CIFAR-10, SVHN, CIFAR-100 and ImageNet. Then, we visualize the histograms of the regularized activation distribution. Finally, we analyze the robustness of our approach to the hyper-parameter selection. 

\subsection{Accuracy Improvement}\label{sec:exp-accuracy-improvement}
\paragraph{Training configuration.} We use fully convolutional VGG-style networks for CIFAR-10 and SVHN, and ResNet for CIFAR-100. All of them use the ADAM optimizer~\cite{kingma2014adam} as suggested by Hubara \textit{et al.}~\cite{hubara2016binarized}. For the BNN trained with distribution loss (BNN-DL), we compute the loss with the activations prior to each binarized activation function ($i.e.$, $Sign$ function that uses $HardTanh$ for gradient computation). Unless noted otherwise, we set the coefficient $k_\epsilon$ to be 1, 0.25 and 0.25 for $L_D$, $L_S$ and $L_M$, respectively, and set $\lambda$ to be 2. To show the statistical significance, all the experiments for CIFAR-10 and SVHN are averaged over five experiments with different parameter initialization seeds. The details of the network structure and training scheme for each dataset is as follows:

\textit{CIFAR-10.}
The network structure can be formulated as: $x$C-$x$C-MP-$2x$C-$2x$C-MP-$4x$C-$4x$C-10C-GP, where $x$C indicates a convolutional layer with $x$ filters, MP and GP indicate max pooling and global pooling layers, respectively. $3\times3$ filter size is used for all the convolutional layers. We vary the $x$ to different values ($\{128, 179, 256, 384\}$) to explore the trade-off between accuracy and energy cost, which are shown in Table~\ref{table:exp-accuracy-improvement} as networks 2-5. We also train a small BNN without the two $4x$C layers for CIFAR-10, which is network 1 in Table~\ref{table:exp-accuracy-improvement}. Each convolutional layer has binarized weights and is followed by a batch normalization layer and a sign activation function. The learning rate schedule follows the code from BNN authors~\cite{BNN_code}. All networks are trained for 200 epochs. 

\textit{SVHN.} 
The network structure for SVHN is the same as CIFAR-10, except that the $x$ is varied from $\{51, 64, 96, 128\}$, shown by networks 6-9 in Table~\ref{table:exp-accuracy-improvement}. The initial learning rate value is 1e-2, and decays by a factor of 10 at epochs 20, 40 and 45. We train 50 epochs in total.

\textit{CIFAR-100.}
We use the \textit{full pre-activation} variant of ResNet~\cite{he2016identity} with 20 layers for CIFAR-100. Prior work has shown the difficulty of training ResNet-based BNNs without scaling layers~\cite{yazdani2018linear}. Since in ResNet-based BNNs the main path activations and residual path activations do not have matching scales, directly adding the two activations will cause difficulty in training. Therefore, we add two batch normalization layers after these two activations to maintain stable activation scales. 
Similar to CIFAR-10 and SVHN, we also vary the number of filters per layer. Details on the network structure are included in Appendix~\ref{appendix:network-structure}. The initial learning rate is set to 1e-4, and reduced by a factor of 3 every 80 epochs. The networks are trained for 300 epochs.

\begin{table*}[t]
\begin{center}
\caption{Accuracy improvement with distribution loss. Network depth is defined as the number of convolutional layers, while the network width is defined as the number of filters in the largest layer. The best results are shown in bold face. \textit{All the accuracy for CIFAR-10 and SVHN is averaged over five experiments with different weight initialization.}}
\label{table:exp-accuracy-improvement}
\begin{tabular}{ccccccc}
\hline
\multirow{2}{*}{Dataset} & \multirow{2}{*}{Network ID} & \multirow{2}{*}{Depth/width} & \multirow{2}{*}{Params storage} & \multirow{2}{*}{Energy cost ($\mu$J)} & \multicolumn{2}{c}{Accuracy (mean $\pm$ std) (\%)} \\
\hhline{~~~~~--}
& & & & & BNN & BNN-DL \\
\hline
\multirow{5}{*}{CIFAR-10} & 1 & 5/256 & 0.4 MB & 0.30 & 80.61 $\pm$ 0.49 & \textbf{83.33 $\pm$ 0.32} \\
& 2 & 7/512 & 0.6 MB & 0.47 & 87.54 $\pm$ 0.38 & \textbf{89.13 $\pm$ 0.23} \\
& 3 & 7/716 & 1.1 MB & 0.93 & 88.99 $\pm$ 0.13 & \textbf{90.28 $\pm$ 0.28} \\
& 4 & 7/1024 & 2.3 MB & 1.89 & 90.09 $\pm$ 0.10 & \textbf{91.01 $\pm$ 0.09} \\
& 5 & 7/1536 & 3.8 MB & 4.23 & 90.68 $\pm$ 0.11 & \textbf{91.56 $\pm$ 0.16} \\
\hline
\multirow{4}{*}{SVHN} & 6 & 7/204 & 0.09 MB & 0.08 & 96.23 $\pm$ 0.15 & \textbf{96.57 $\pm$ 0.12} \\
& 7 & 7/256 & 0.15 MB & 0.12 & 96.53 $\pm$ 0.11 & \textbf{96.95 $\pm$ 0.10} \\
& 8 & 7/384 & 0.3 MB & 0.27 & 97.15 $\pm$ 0.15 & \textbf{97.34 $\pm$ 0.05} \\
& 9 & 7/512 & 0.6 MB & 0.47 & 97.34 $\pm$ 0.07 & \textbf{97.51 $\pm$ 0.03} \\
\hline
\multirow{3}{*}{CIFAR-100} & 10 & 20/1024 & 5.6 MB & 53.7 & 60.40 & \textbf{68.17} \\
& 11 & 20/1536 & 12.6 MB & 120.9 & 64.57 & \textbf{71.53} \\
& 12 & 20/2048 & 22.3 MB & 215.0 & 66.07 & \textbf{73.42} \\
\hline
\end{tabular}
\vspace{-9pt}
\end{center}
\end{table*}

\begin{figure}[t]
  \centering
  \includegraphics[width=0.47\textwidth]{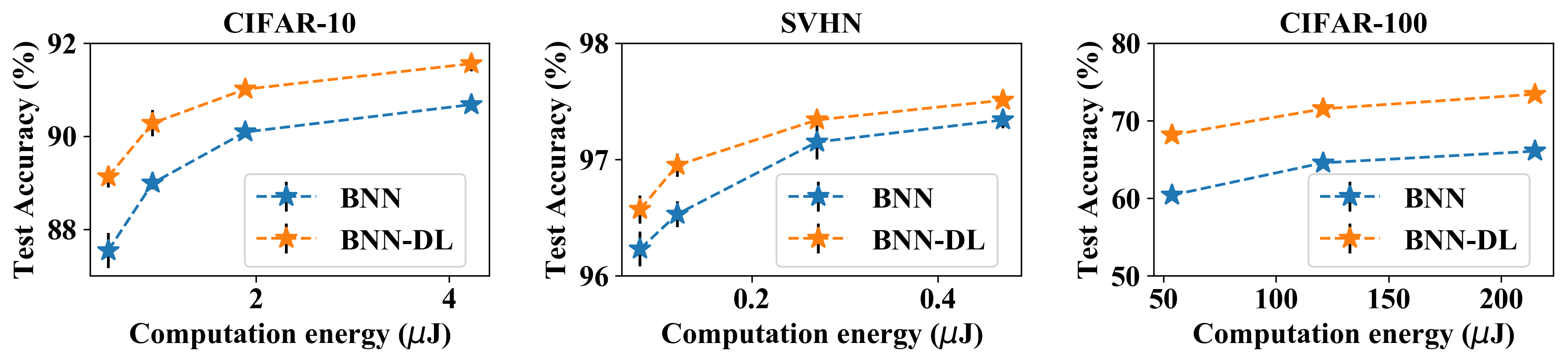}
  \caption{Accuracy and energy Pareto-optimal curve for CIFAR-10, SVHN and CIFAR-100. The error bars for CIFAR-10 and SVHN show the standard deviation of testing accuracy.}\label{fig:exp-pareto}
  \vspace{-15pt}
\end{figure}

\vspace{-12pt}
\paragraph{Results on CIFAR-10, SVHN and CIFAR-100.} 
As shown in Table~\ref{table:exp-accuracy-improvement}, the accuracy for BNN-DL is consistently higher than the baseline BNN. The accuracy gap between BNN and BNN-DL is generally larger than their standard deviations. Using t-test, the p-values for all the network 1-9 are smaller than 0.005, which demonstrates the statistical significance of our improvements. In addition to accuracy results, we also show the computational energy cost for each network, obtained by summing up the energy of each operation for the inference of a single input image. Note that this cost excludes the energy of memory accesses, which is the same for BNN and BNN-DL in the inference phase. We also visualize the trade-off between accuracy and energy cost in Fig.~\ref{fig:exp-pareto}. In most cases, the BNN-DL with a smaller model size can achieve the same or higher accuracy than the BNN with a larger size. 

\begin{figure}[t]
  \centering
  \vspace{-12pt}
  \subfloat[CIFAR-10, network-1]{\includegraphics[width=0.24\textwidth]{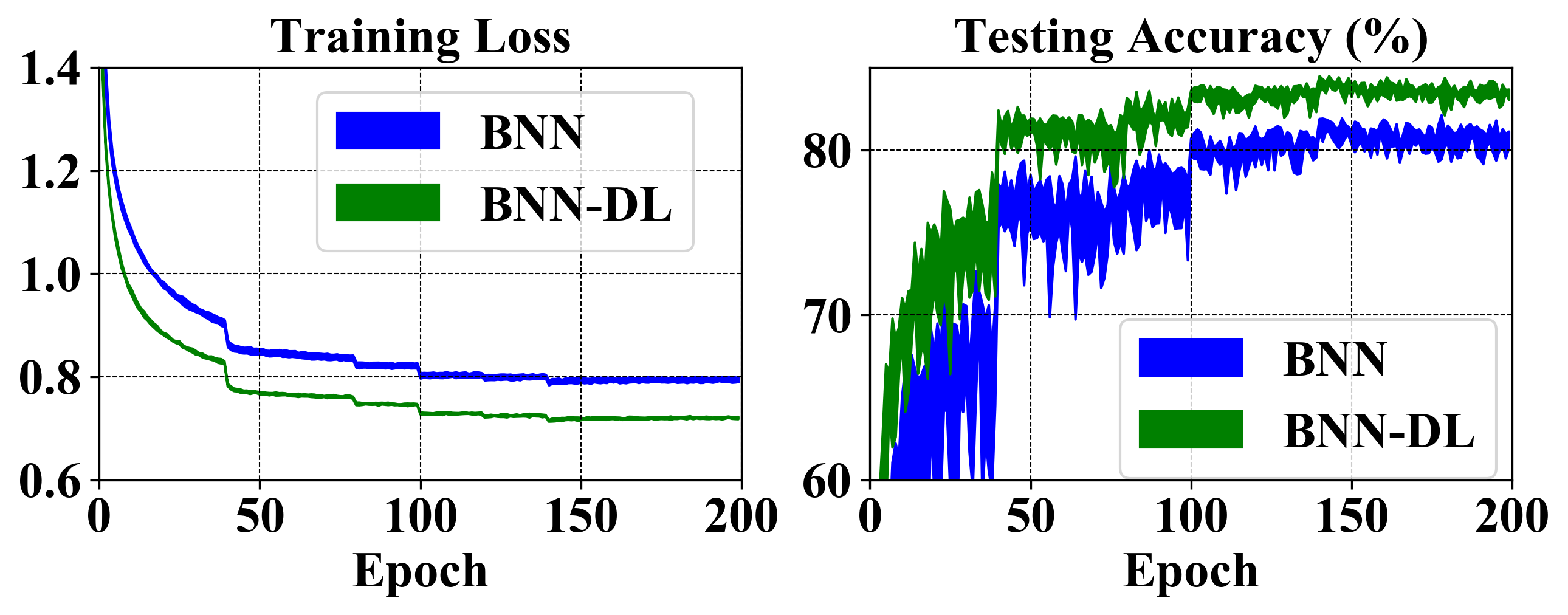}\label{fig:exp-curve-1}}
  \subfloat[CIFAR-10, network-5]{\includegraphics[width=0.24\textwidth]{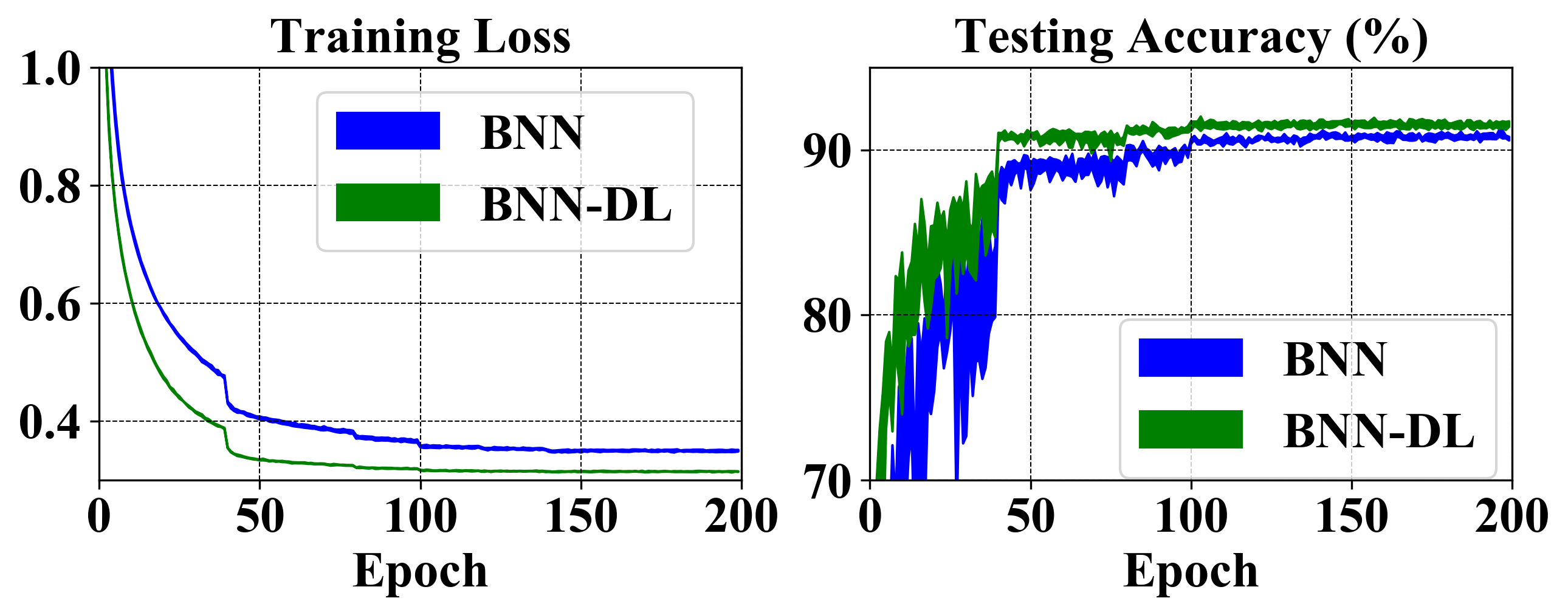}\label{fig:exp-curve-2}}\\
  \subfloat[SVHN, network-6]{\includegraphics[width=0.24\textwidth]{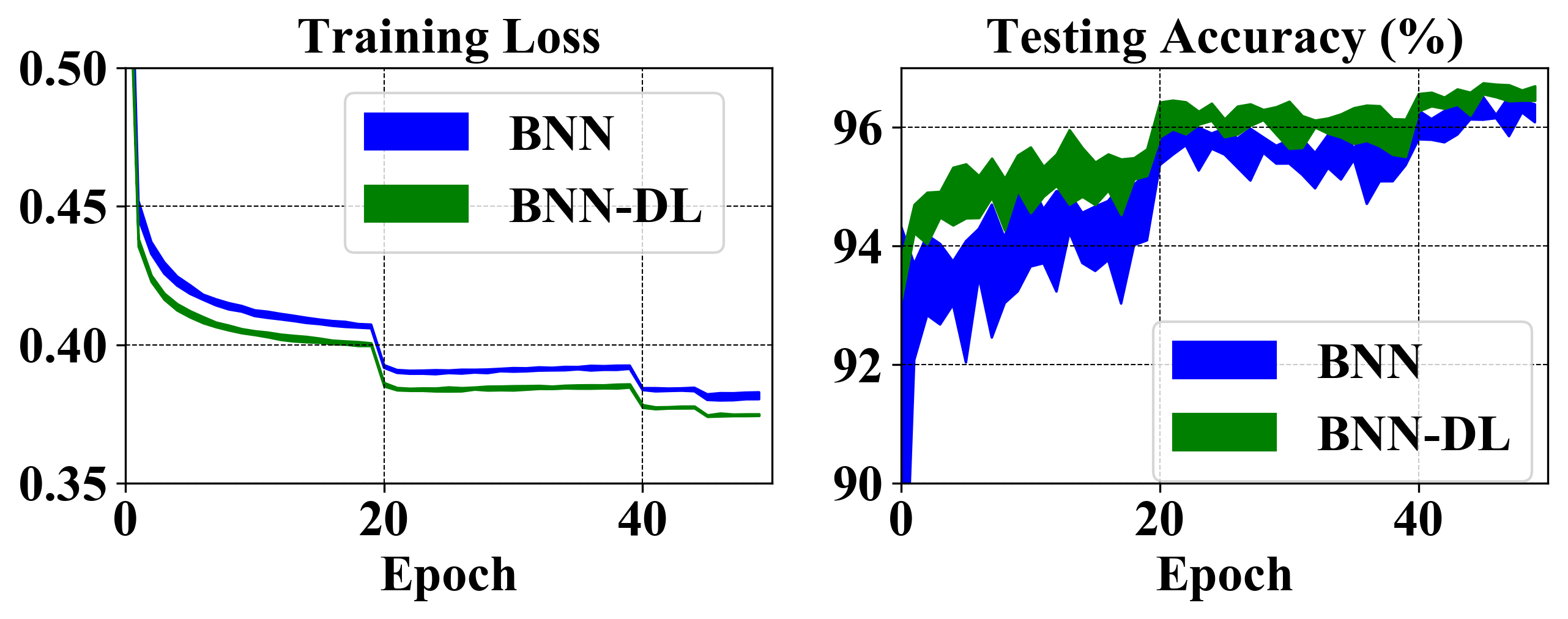}\label{fig:exp-curve-3}}
  \subfloat[SVHN, network-9]{\includegraphics[width=0.24\textwidth]{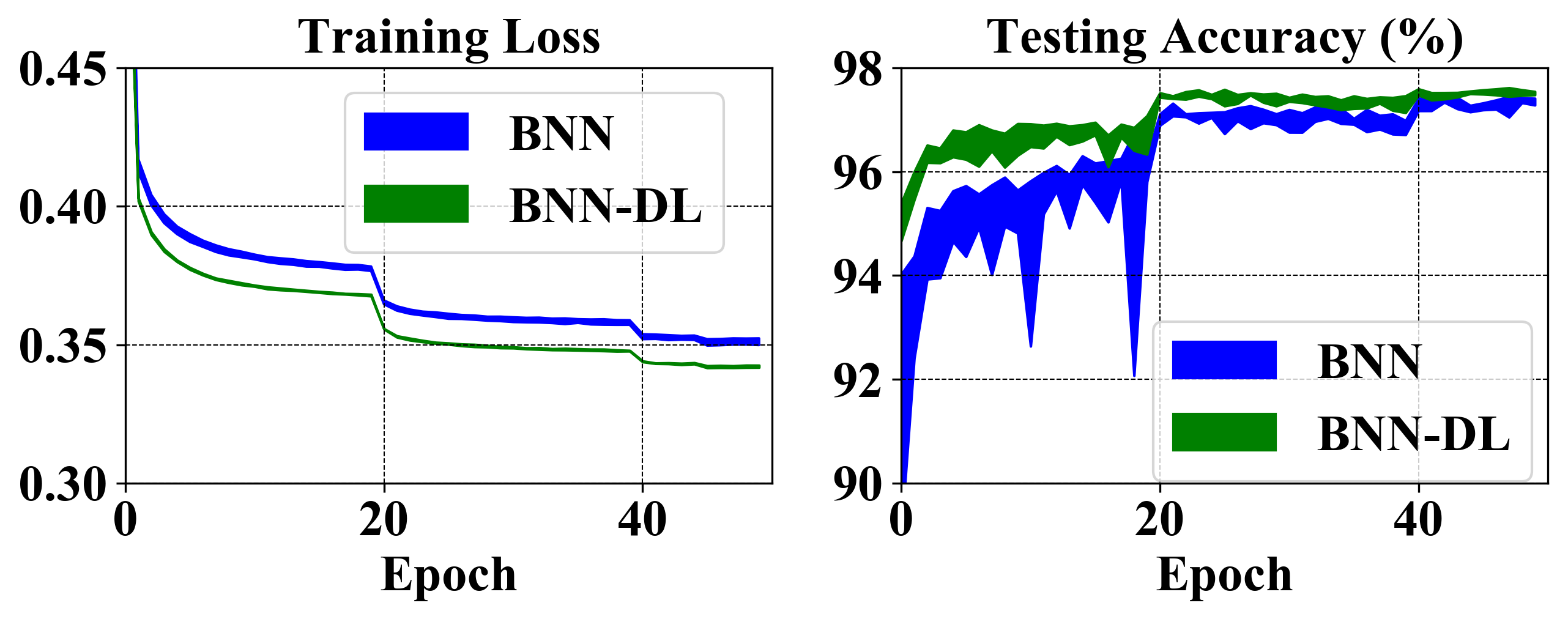}\label{fig:exp-curve-4}}\\
  \subfloat[CIFAR-100, network-10]{\includegraphics[width=0.24\textwidth]{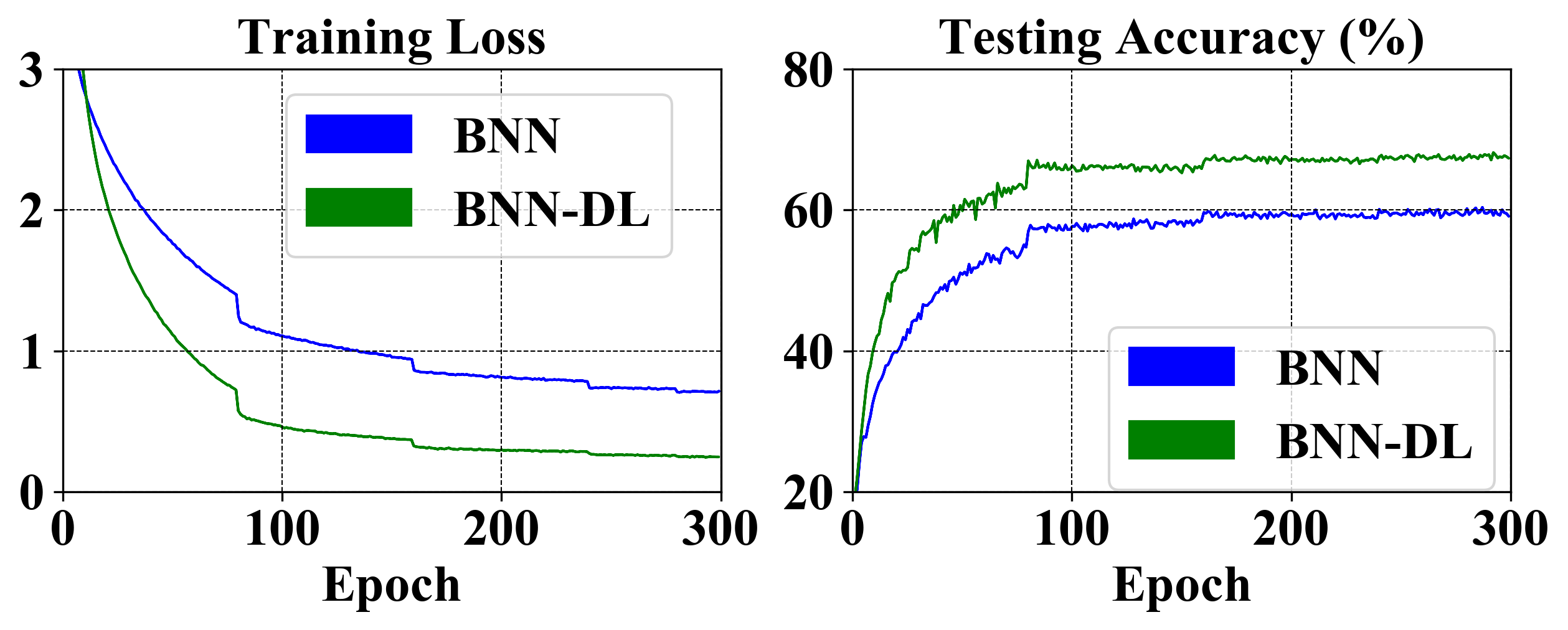}\label{fig:exp-curve-5}}
  \subfloat[CIFAR-100, network-12]{\includegraphics[width=0.24\textwidth]{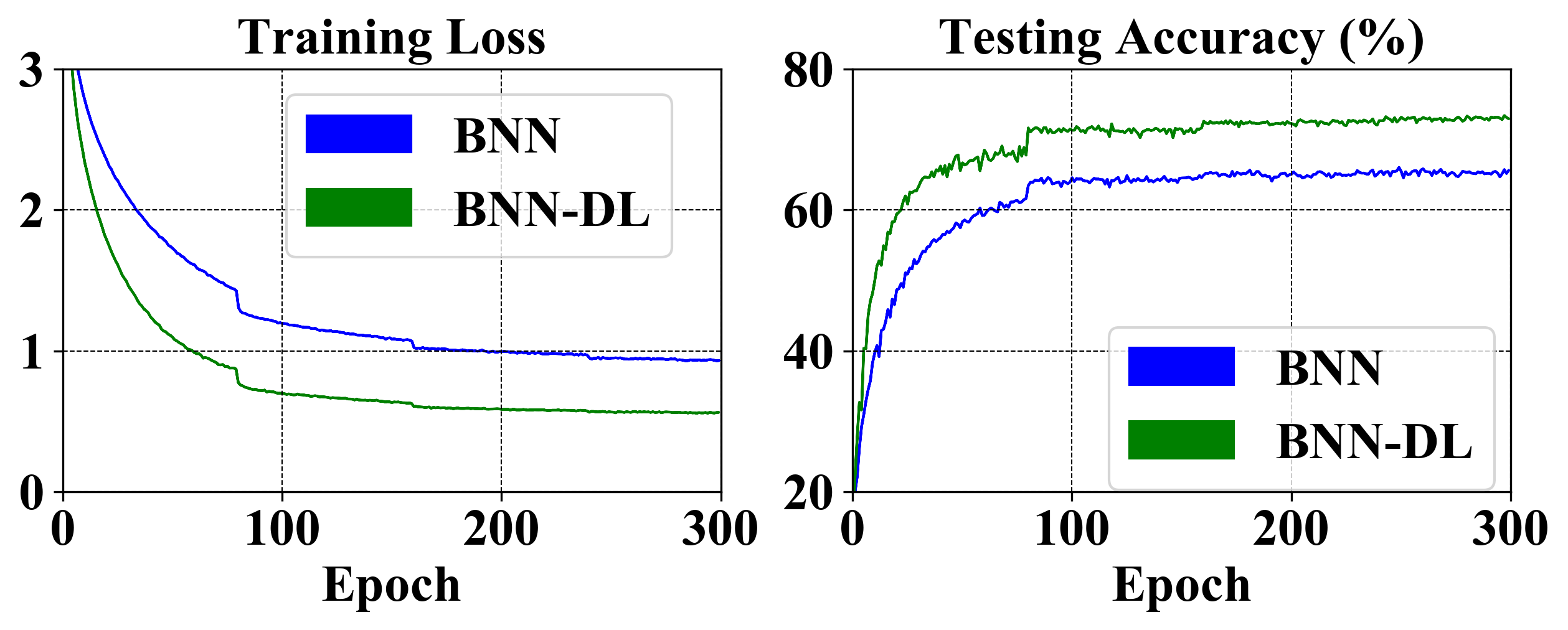}\label{fig:exp-curve-6}}\\
  \caption{Training loss and testing accuracy curves for different networks with or without distribution loss. The widths of the curves for CIFAR-10 and SVHN are 2 standard deviation ranges.}\label{fig:exp-curve}
  \vspace{-15pt}
\end{figure}

The use of the distribution loss improves the testing accuracy mostly because it regularizes the activation and gradient flow in the training phase, so that the networks can better fit the dataset. As shown in Fig.~\ref{fig:exp-curve}, the training loss for BNN-DL is consistently lower than the BNN baseline after a few epochs. For most of the experiments, distribution loss is found to converge to a very small number (\textit{e.g.}, $1/10000$ of the initial value) in the first few epochs. This indicates that the network can be easily regularized by the distribution loss, which then improves the rest of the training process.

\begin{table*}
\begin{tabular}{cc}
    \begin{minipage}{.65\linewidth}
    \begin{center}
    \caption{Comparison with prior art using 1-bit weights and activations, in terms of accuracy and computation energy on different datasets. The best results are shown in bold face.}
    \vspace{-9pt}
    \label{table:comparison-prior}
        \begin{tabular}{ccccc}
        \hline
        Dataset & Model & Pure-logical & Energy cost & Accuracy \\
        \hline
        \multirow{4}{*}{CIFAR-10} & BNN~\cite{hubara2016binarized} & Yes & 1$\times$ & 87.13\% \\
        & XNOR-Net~\cite{rastegariECCV16} & No & 4.5$\times$ & 87.38\% \\
        & LAB~\cite{hou2017lossaware} & No & 4.5$\times$ & 87.72\% \\
        & \textbf{BNN-DL} & \textbf{Yes} & $1\times$ & \textbf{89.90\%} \\
        \hline
        \multirow{4}{*}{SVHN} & BNN~\cite{hubara2016binarized} & Yes & 1$\times$ & 96.50\% \\
        & XNOR-Net~\cite{rastegariECCV16} & No & 4.5$\times$ & 96.57\% \\
        & LAB~\cite{hou2017lossaware} & No & 4.5$\times$ & 96.64\% \\
        & \textbf{BNN-DL} & \textbf{Yes} & 1$\times$ & \textbf{97.23\%} \\
        \hline
        \multirow{3}{*}{CIFAR-100} & BNN~\cite{hubara2016binarized} & No & 1$\times$ & 60.40\% \\
        & DQ-2bit~\cite{polino2018model} & No & - & 49.32\% \\
        & \textbf{BNN-DL} & \textbf{No} & 1$\times$ & \textbf{68.17\%} \\
        \hline
        \end{tabular}
    \end{center}
    \end{minipage} &

    \begin{minipage}{.3\linewidth}
    \begin{center}
    \caption{Robustness to the selection of optimizer, learning rate, and network structure. CIFAR-10 is used for illustrating the results.}
    \label{table:robust-optimizer}
    \begin{tabular}{ccc}
    \hline
    & BNN & BNN-DL \\
    \hline
    Momentum & 66.02\% & 89.37\% \\
    Nesterov & 68.66\% & 89.22\% \\
    Adam & 88.12\% & 89.62\% \\
    RMSprop & 87.39\% & 90.24\% \\
    \hline
    lr$_{\text{init}}$=1e-1 & 82.19\% & 89.60\% \\
    lr$_{\text{init}}$=5e-3 & 88.12\% & 89.62\% \\
    lr$_{\text{init}}$=2e-4 & 85.62\% & 88.73\% \\
    \hline
    VGG & 88.12\% & 89.62\% \\
    ResNet-18 & 85.71\% & 90.47\% \\
    \hline
    \end{tabular}
    \end{center}
    \end{minipage} 
\end{tabular}
\end{table*}

\vspace{-6pt}
\paragraph{Comparison with prior art.}
We also compare our results with prior work on binarized networks as shown in Table~\ref{table:comparison-prior}. For CIFAR-10 and SVHN, we follow the same network configuration by Hou \textit{et al.}, and also split the dataset into training, validation and testing sets as they do~\cite{hou2017lossaware}. Table~\ref{table:comparison-prior} shows that by just applying the distribution loss when training BNNs can achieve higher accuracy than the baseline BNN~\cite{hubara2016binarized}, XNOR-Net~\cite{rastegariECCV16} and LAB~\cite{hou2017lossaware}. We also show the normalized energy cost for the models. Since XNOR-Net and LAB use scaling factors for the weights and activations, which introduces the need for full-precision operations, XNOR-Net and LAB require $4.5\times$ energy cost than BNN and BNN-DL. We use 16-bit fixed-point multipliers and adders instead of 32-bit floating-point operators to estimate the energy cost of these full-precision operations because prior quantization work shows that 16-bit fixed-point operation is generally enough for maintaining accuracy~\cite{gupta2015deep}. For CIFAR-100, the closest work that reports 1-bit weights and low-bit activations is by Polino \textit{et al.}~\cite{polino2018model}, where they use a 7.9MB ResNet with 2-bit activations for CIFAR-100, which presumably has larger energy cost than our 5.6MB model with 1-bit activations, and our results on accuracy surpass theirs by a large margin.

\vspace{-12pt}
\paragraph{Results on ImageNet.}
Having shown the effectiveness of the distribution loss on small datasets, we extend our analysis to a larger image dataset - ImageNet ILSVRC-2012~\cite{russakovsky2015imagenet}. We consider AlexNet, which is the most commonly adopted network in prior art on binarized DNNs~\cite{hubara2016binarized,rastegariECCV16,zhou2016dorefa,tang2017train,mishra2018wrpn}. We compare our BNN-DL with the baseline BNN~\cite{hubara2016binarized}, XNOR-Net~\cite{rastegariECCV16}, DoReFa-Net~\cite{zhou2016dorefa}, Compact Net~\cite{tang2017train}, and WRPN~\cite{mishra2018wrpn}. The BNN uses binarized weights for the whole network~\cite{hubara2016binarized}, while XNOR-Net and DoReFa-Net keep the first convolutional layer and last fully-connected layer with full-precision weights~\cite{rastegariECCV16,zhou2016dorefa}. Compact Net uses full-precision weights for the first layer but binarizes the last layer, and uses 2 bits for the activations~\cite{tang2017train}. WRPN doubles the filter number of XNOR-Net, and uses full-precision weights for both the first and last layers~\cite{mishra2018wrpn}. Also, BNN uses 64 and 192 filters while the other networks use 96 and 256 filters (or doubling these numbers as WRPN does) for the first two convolutional layers. We train our BNN-DL using the same settings as prior work, except that we use 1-bit activations instead of 2-bit when comparing with Compact Net. The learning rate policy follows prior implementations~\cite{BNN_code}, but starts from 0.01. As shown in Table~\ref{table:exp-imagenet}, BNN-DL consistently outperforms the accuracy of the baseline models. All baseline models except BNN use scaling factors to approximate activations while we keep them binarized. Therefore, our model also has lower energy cost than the prior models. In addition, we highlight that our BNN-DL can outperform Compact Net though we use fewer bits for activations. 

\begin{table}[t]
\begin{center}
\caption{Comparison with prior art on ImageNet with AlexNet-based topology. We use the same model structure as prior work, except that Compact Net uses 2 bits for activations while we only use 1 bit. Training with distribution loss outperforms prior work consistently.}
\label{table:exp-imagenet}
\vspace{-6pt}
\begin{tabular}{c|cc|cc}
\hline
\multirow{2}{*}{Model} & \multicolumn{2}{c|}{Baseline} & \multicolumn{2}{c}{Ours} \\
\hhline{~----}
& Top-1 & Top-5 & Top-1 & Top-5 \\
\hline
BNN~\cite{hubara2016binarized} & 36.1\% & 60.1\% & 41.3\% & 65.8\% \\
XNOR-Net~\cite{rastegariECCV16} & 44.2\% & 69.2\% & 47.8\% & 71.5\% \\
DoReFa-Net~\cite{zhou2016dorefa} & 43.5\% & - & 47.8\% & 71.5\% \\
Compact Net~\cite{tang2017train} & 46.6\% & 71.1\% & 47.6\% & 71.9\% \\
WRPN~\cite{mishra2018wrpn} & 48.3\% & - & 53.8\% & 77.0\% \\
\hline 
\end{tabular}
\vspace{-20pt}
\end{center}
\end{table}

\subsection{Regularized Activation Distribution}\label{sec:exp-regularized}
\begin{figure}[t]
  \centering
  \subfloat[Baseline BNN]{\includegraphics[width=0.47\textwidth]{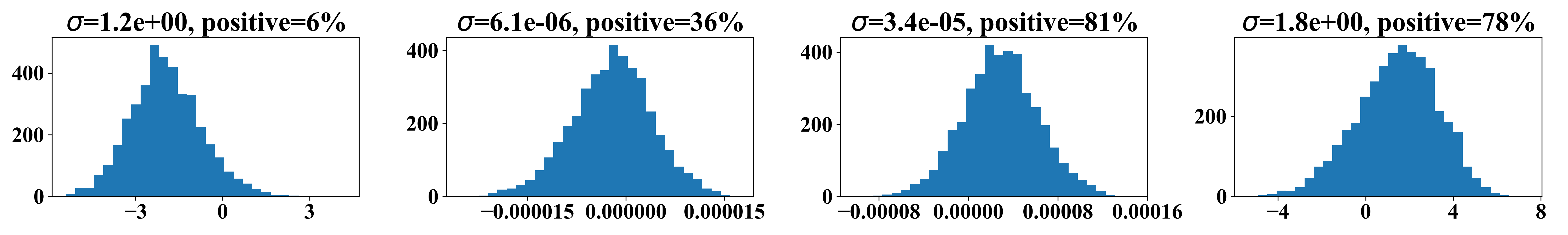}\label{fig:exp-dist-1}}\\
  \vspace{-9pt}
  \subfloat[BNN trained with distribution loss]{\includegraphics[width=0.47\textwidth]{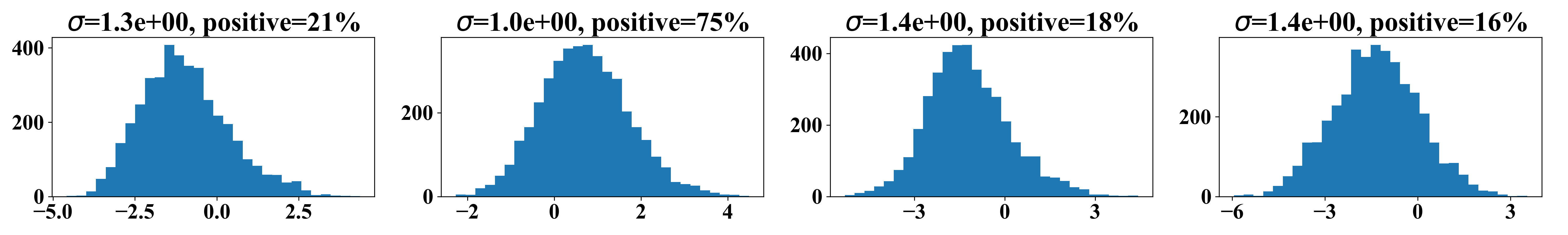}\label{fig:exp-dist-2}}\\
  \vspace{-6pt}
  \caption{Activation distribution for BNN trained (a) without or (b) with distribution loss. Each histogram refers to the activations of one channel. In (a), the channel in the left histogram shows a generation problem, the middle two show gradient mismatch, and the right one shows saturation problem. $\sigma$ is standard deviation, and ``positive" refers to the ratio of positive activations.}\label{fig:exp-dist}
  \vspace{-6pt}
\end{figure}

To show the regularization effect of the distribution loss, we plot the distribution of the pre-activations for the baseline BNN and for our proposed BNN-DL. More specifically, we conduct inference for network 2 on CIFAR-10, and extract the (floating-point) activations right after the batch normalization layer prior to the binarized activation function of the fourth convolutional layer with 256 filters. Therefore, for each of the 256 output channels, we get its values across the whole dataset. Then, for illustration purposes, we select four channels from baseline BNN and our proposed BNN-DL, respectively, and plot the histogram of these per-channel values, as shown in Fig.~\ref{fig:exp-dist}. The four channels' activation distributions for the baseline BNN are picked to show the degeneration, gradient mismatch, and saturation problems, while the distributions for BNN-DL are randomly selected. From Fig.~\ref{fig:exp-dist-1} we can see that the good weight initialization strategy~\cite{glorot2010understanding} and batch normalization~\cite{ioffe2015batch} adopted for BNNs do not solve the distribution problems. 

\begin{figure}[t]
  \centering
  \includegraphics[width=0.47\textwidth]{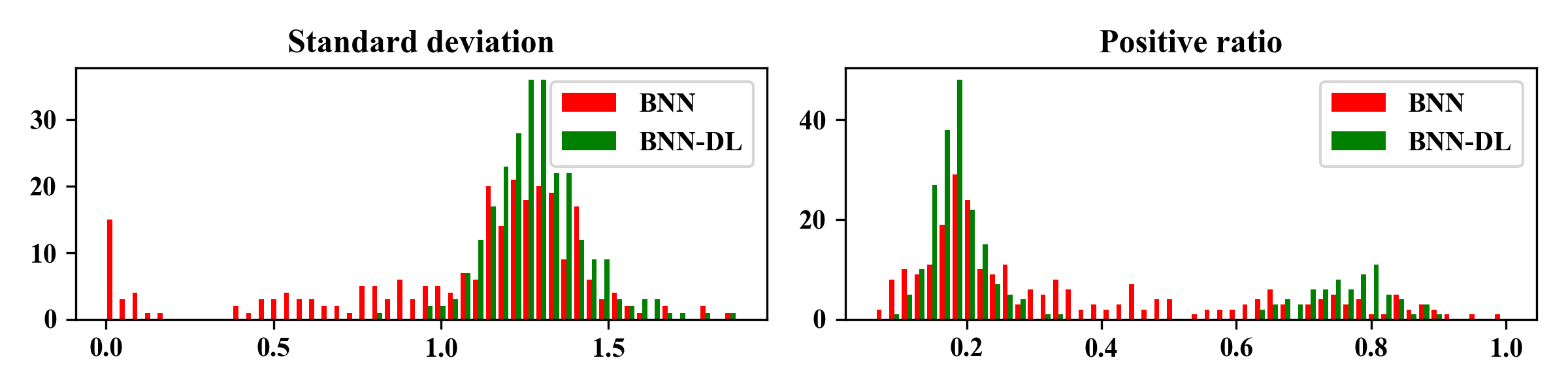}
  \vspace{-9pt}
  \caption{Histogram of standard deviation and positive ratio of per-channel activations. 
  }\label{fig:exp-std-pos-dist}
  \vspace{-12pt}
\end{figure}

To show that BNN-DL alleviates these challenges, we compute the standard deviation of activations for each of these 256 channels, as well as their positive ratio, which is the proportion of positive values. As shown in Fig.~\ref{fig:exp-std-pos-dist}, the standard deviation of BNN-DL is more regularized and centralized than that of BNN. The channel with very small standard deviation like the middle two histograms in Fig.~\ref{fig:exp-dist-1} is rarely seen in BNN-DL, while BNN has a long tail in the area of small standard deviations. This indicates that without explicit regularization, the scale factors of batch normalization layer could shrink to very small values, causing the gradient mismatch problem. 
From Fig.~\ref{fig:exp-std-pos-dist}, we can also observe that the positive ratio of BNN has more extreme values (\textit{i.e.}, those close to 0 or 1) than BNN-DL. This indicates that the degeneration problem is reduced by distribution loss. Interestingly, we can see that the positive ratio of BNN-DL also deviates away from 0.5. We conjecture that this is because the activations centered at 0 are more prone to gradient mismatch, and thus, be penalized by $L_M$. 

\subsection{Robustness to Hyper-parameter Selection}
\vspace{-6pt}
Another benefit of the distribution loss is its robustness to the selection of the training hyper-parameters. Prior work~\cite{alizadeh2018a} has shown that the accuracy of BNNs is sensitive to the training optimizer. We observe the same phenomenon by training BNNs with different optimizers including SGD with momentum, SGD with Nesterov~\cite{sutskever2013importance}, Adam~\cite{kingma2014adam} and RMSprop~\cite{tieleman2012lecture}. However, when training BNN with distribution loss, these optimizers can be consistently improved, as shown in Table~\ref{table:robust-optimizer}. We use CIFAR-10 for the experiments in this subsection. We use the same weight decay and learning rate schedule as Zagoruyko \textit{et al.}~\cite{zagoruyko2016wide} for Momentum and Nesterov, and change the initial learning rate to 1e-4 for RMSprop. We use the same setting as Hubara \textit{et al.}~\cite{BNN_code} for Adam. Each model is trained for 200 epochs, and the best testing accuracy is reported. 
Then, we vary the learning rate schedule of Hubara \textit{et al.}'s implementation~\cite{BNN_code} by scaling the learning rate at each epoch by a constant. Table~\ref{table:robust-optimizer} shows that BNN-DL is more robust to the selection of learning rate values. 
Furthermore, we show that the BNN-DL can work well for both VGG-style networks with stacked convolutional layers and ResNet-18 which includes skip connections. The VGG-style network uses the network 2 in Table~\ref{table:exp-accuracy-improvement}. The ResNet-18 structure uses the pre-activation variant~\cite{he2016identity} with added batch normalization layers as described in Sec.~\ref{sec:exp-accuracy-improvement}. Since BNN has non-regularized activations, maintaining the activation flow in the training process requires more careful picking of hyper-parameter values. However, the distribution loss applies regularization to the activations, making the network easier to train, and therefore reduces the sensitivity to hyper-parameter selection. 

We also show that the distribution loss is robust to the selection of the introduced hyper-parameter, $\lambda$ coefficient, which indicates the regularization level of the distribution loss. As shown in Table~\ref{table:robustness-coeff}, by varying $\lambda$ from 0.2 to 2000, the accuracy for BNN-DL is consistently higher than the baseline BNN. As mentioned in Sec.~\ref{sec:exp-accuracy-improvement}, the distribution loss quickly decays to a small magnitude in the first few epochs, and we find that this holds for a wide range of $\lambda$. The robustness analysis indicates that the distribution loss is a handy tool to regularize activations, without the need of much hyper-parameter tuning.

\begin{table}[t]
\begin{center}
\caption{Accuracy for BNN-DL on CIFAR-10 with varied regularization levels. 
$\lambda=0$ indicates the baseline BNN.
}
\vspace{-9pt}
\label{table:robustness-coeff}
\begin{tabular}{C{1.3cm}C{0.7cm}C{0.7cm}C{0.7cm}C{0.7cm}C{0.7cm}C{0.7cm}}
\hline
$\lambda$ & 0 & 0.2 & 2 & 20 & 200 & 2000 \\
\hline
Acc. (\%) & 87.39 & 90.12 & 90.16 & 90.18 & 90.61 & 90.20 \\
\hline
\end{tabular}
\vspace{-18pt}
\end{center}
\end{table}



\vspace{-6pt}
\section{Conclusion}
\vspace{-9pt}
In this paper, we tackle the difficulty of training BNNs with 1-bit weights and 1-bit activations. The difficulty arises from the unregularized activation flow that may cause degeneration, saturation and gradient mismatch problems. We propose a framework to embed this insight into the loss function by formulating our hypothesis, adjusting regularization and enabling differentiability, and thus, explicitly penalizing the activation distributions that may lead to the training problems. Our experiments show that BNNs trained with the proposed distribution loss have regularized activation distribution, and consistently outperform the baseline BNNs. The proposed approach can significantly improve the accuracy of the state-of-the-art networks using 1-bit weights and activations for AlexNet on ImageNet dataset. In addition, this approach is robust to the selection of training hyper-parameters including learning rate and optimizer. These results show that distribution loss can generally benefit the training of binarized networks which enable latency and energy efficient inference on mobile devices.

\vspace{-12pt}
\section*{Acknowledgement}
\vspace{-9pt}
This research was supported in part by NSF CCF Grant No. 1815899, and NSF award number ACI-1445606 at the Pittsburgh Supercomputing Center (PSC).

{\small
\bibliographystyle{ieee}
\bibliography{cvpr2019}
}

\clearpage
\newpage
\section{Appendix}

\subsection{Energy cost for different types of binarization}\label{appendix:energy-cost}
We adopt a convolutional layer from VGG-16 for ImageNet to estimate the computation energy cost in Table~\ref{table:layer-energy}. In the chose layer, both input and output channels are 256; both input and output feature maps are 56x56; the kernels are 3x3 with stride 1. 
To estimate the computational energy consumption for BNN~\cite{hubara2016binarized}, XNOR-Net~\cite{rastegariECCV16} and ABC-Net~\cite{lin2017towards}, we first compute the number of XNORs, counts, fixed-point multiplications and additions for each of the binarization methods, and then add the energy consumption for all the operations together. For BNN and XNOR-Net, the number of XNORs is $C_{in}C_{out}HWK_hK_w$, where $C_{in}$, $C_{out}$, $H$, $W$, $K_h$, $K_w$ are input channels, output channels, output feature map height and width, and kernel height and width, respectively. For ABC-Net the number of XNORs is $MNC_{in}C_{out}HWK_hK_w$ where $M$ and $N$ are the number of bases for weights and activations, respectively. In addition to the XNOR operations, BNN also needs counts and comparators, where the number of counts is roughtly the same as XNORs, and the number of comparators is $C_{out}HW$. XNOR-Net also needs $2C_{out}HW$ fixed-point multiplications and $C_{out}HWK_hK_w$ additions, as shown in Fig.~\ref{fig:xnor_block}, since the multiplications within the convolution between $A^{avg}$ and $k$ can be combined with the following scaling operation. ABC-Net needs approximately $MNC_{out}HW$ multiplications and additions as shown in Fig.~\ref{fig:abc_block}.

\subsection{XNOR-Net and ABC-Net blocks}
\begin{figure}[h]
  \centering
  \includegraphics[width=0.47\textwidth]{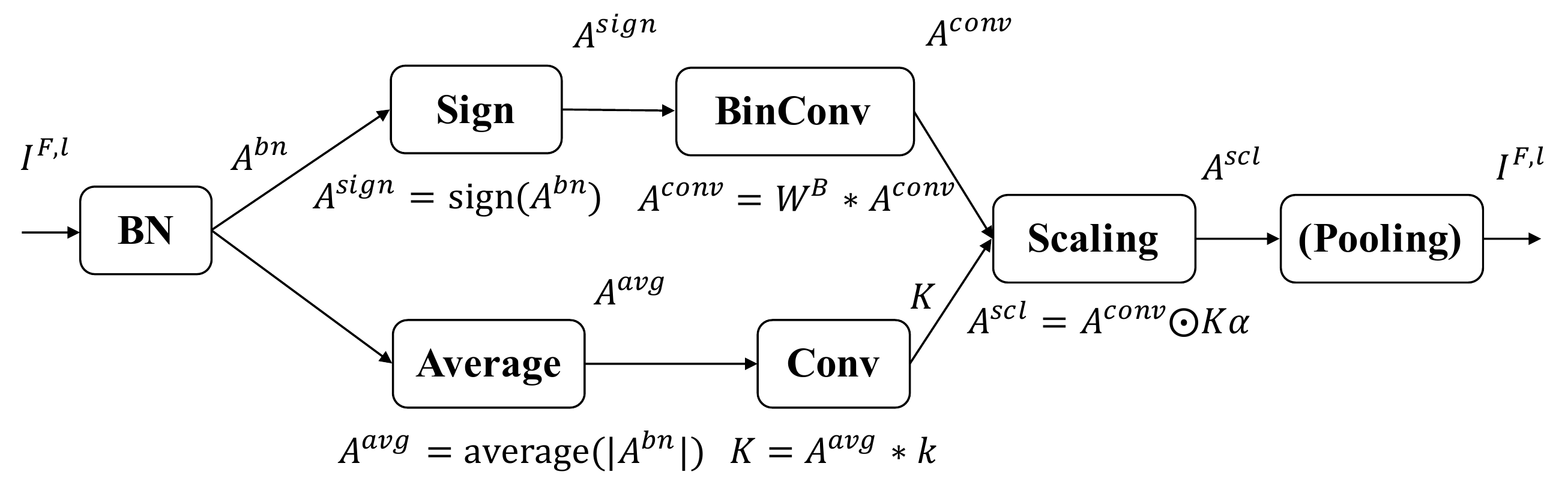}
  \caption{Basic block for XNOR-Net~\cite{rastegariECCV16}.}\label{fig:xnor_block}
\end{figure}

\begin{figure}[h]
  \centering
  \includegraphics[width=0.47\textwidth]{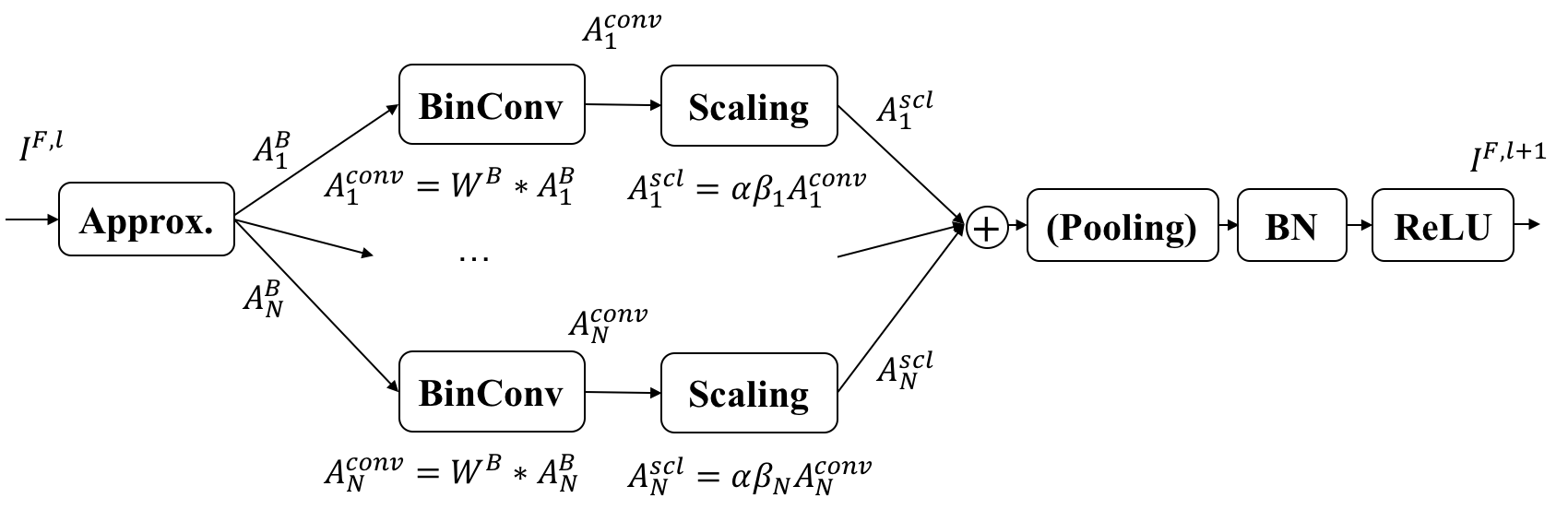}
  \caption{Basic block for ABC-Net~\cite{lin2017towards}.}\label{fig:abc_block}
\end{figure}

The basic blocks of XNOR-Net and ABC-Net are shown in Fig.~\ref{fig:xnor_block} and Fig.~\ref{fig:abc_block}. 

\subsection{Network structure for CIFAR-100}\label{appendix:network-structure}
\begin{figure}[h]
  \centering
  \includegraphics[width=0.47\textwidth]{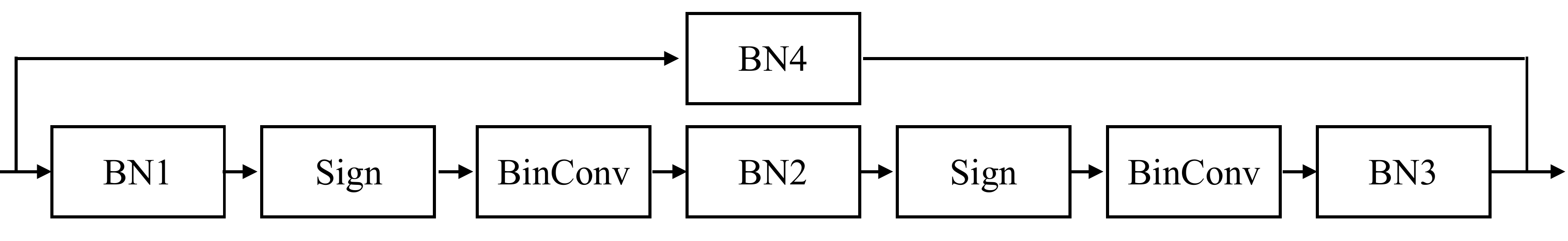}
  \caption{Basic block for ResNet-based BNN~\cite{lin2017towards}.}\label{fig:resnet_bnn_block}
\end{figure}

The basic block for ResNet-based BNN is shown in Fig.~\ref{fig:resnet_bnn_block}. Compared to the full-precision resnet, we add two convolutional layers, BN3 and BN4, to maintain a stable activation flow. For BNN-DL, the distribution loss is still applied to the activations prior to two sign activation functions. 

The network structure for CIFAR-100 is: $x$C-$x$B-$x$B-$2x$B-$2x$B-$4x$B-$4x$B-$8x$B-$8x$B-GP-100L, where $x$C indicates a convolutional layer with $x$ filters, $x$B indicates a basic block with $x$ filters for each convolutional layers, GP means global pooling, and $x$L means a linear layer with $x$ output neurons. All the convolutional layers use 3$\times$3 filter sizes. The first convolutional layers within the 2nd, 3rd and 4th blocks use stride 2 to reduce the feature map sizes, while the other convolutional layers use stride 1. We vary $x$ from $\{128, 192, 256\}$. 

\end{document}